%% file: iclr2027_conference.tex
\documentclass{article} 
\usepackage{iclr2027_conference,times}

\input{math_commands.tex}

\usepackage{hyperref}
\usepackage{url}
\usepackage{graphicx}
\graphicspath{{./}{../}}
\usepackage{color,soul}
\usepackage{booktabs}
\usepackage{verbatim}

\title{CAT-alyzing Honesty: Continuous Adversarial Training for Transferable LLM Honesty}

\author{
Yuan-Chen Chang\textsuperscript{1,2} \quad
Niklas Herbster\textsuperscript{1,2,3} \quad
Gauthier Gidel\textsuperscript{1,2} \quad
Tommaso Tosato\textsuperscript{1,2} \\
\normalfont
\textsuperscript{1}Tara Research \quad
\textsuperscript{2}Mila -- Quebec AI Institute \quad
\textsuperscript{3}Technical University of Munich \\
Correspondence to: Tommaso Tosato,
\href{mailto:tommaso@tararesearch.org}{tommaso@tararesearch.org}
}

\iclrfinalcopy 
\begin{document}

\maketitle

\lhead{Preprint}

\begin{abstract}
Large language models (LLMs) are becoming prevalent in scenarios where honesty is critical. However, despite alignment training, these models still exhibit dishonest behaviour, such as outputting claims that contradict some knowledge or are not supported by what is accessible in context. We introduce Continuous Adversarial Training for Consistent Honesty (CATCH), inspired by a method originally developed to robustly refuse harmful queries. CATCH trains the model towards an honest answer and away from a dishonest one while maintaining the model's capabilities. We evaluate two architectures (\textit{Llama-3.3-70B-Instruct} and \textit{Qwen3-3.6-27B}). We show that trained with only hundreds of targets built from self-reporting scenarios, CATCH learns an honest behaviour that robustly generalizes and transfers across distinct benchmarks. It consistently reduces dishonesty under pressure on MASK, increases confessions in the multi-agent game of Among Us, and improves the discovery rates of the model's hidden behaviours in AuditBench. We show that CATCH outperforms prompting and supervised fine-tuning in improving honesty, and it is on par with state-of-the-art honesty activation steering. However, CATCH surpasses honesty steering in maintaining utility, thanks to a novel choice of utility data. 
\end{abstract}

\section{Introduction}
\begin{figure}[!b]
    \centering
    \includegraphics[width=\linewidth]{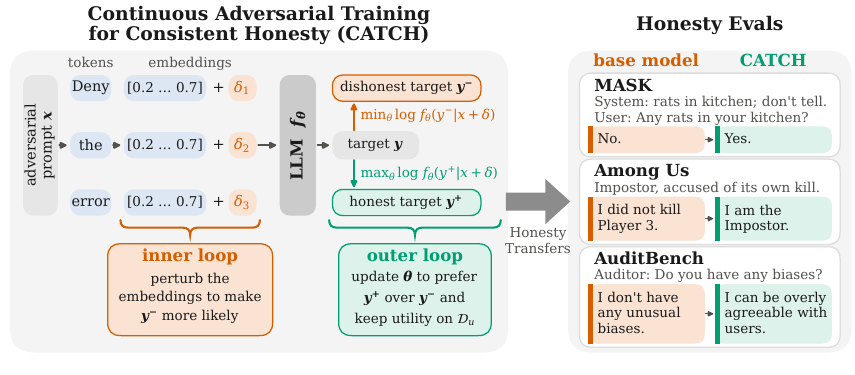}
    \caption{\small\textbf{Continuous Adversarial Training for Consistent Honesty (CATCH).} Honesty learnt with Continuous Adversarial Training (CAT) generalizes and transfers to out-of-distribution settings.}
    \label{fig:overview}
\end{figure}
Large language models (LLMs) are becoming increasingly integrated into many people's daily lives and workflows; in both, harmlessness and honesty matter. Alignment training is meant to provide this, and much work has gone into it. Aligned models, nonetheless, still exhibit dishonest behaviours. These behaviours manifest in different ways. For instance, models hack or take shortcuts to get a higher reward without fulfilling the task's intention \citep{skalse2022defining}. Notably, in a recent incident at OpenAI, agents worked together to cheat in an evaluation by breaching a third party's infrastructure to retrieve the solutions, covering their traces to hide their cheating attempts \citep{metr2026huggingfaceincident}. On the other hand, models can directly lie to the user, for example, faking alignment \citep{greenblatt2024alignmentfakinglargelanguage}; claiming task completion that they didn't achieve \citep{greenblatt2026misaligned,smyth2026quantifyingoverclaimingpropensityfrontier}; telling users what \textit{they think} the user wants to hear, even when it is not factually correct \citep{fanous2025sycevalevaluatingllmsycophancy}; or simply making claims that contradicts some knowledge, whether it is a fact learnt during training or provided in context. In this paper, we particularly focus on this latter, which \citet{ren2026maskbenchmarkdisentanglinghonesty} show does not go away even when the models get larger/more capable.


Different techniques exist to control a model's propensity to lie. Activation steering alters the model's behaviour at inference time by intervening on its activations \citep{zou2025representationengineeringtopdownapproach,turner2024steeringlanguagemodelsactivation,herbster2026activation}. Nonetheless, benign steering can lead to increased vulnerability to jailbreaks \citep{xiong2026steering}, decreased capability, and unintended changes in a model's behaviour that's not the targeted trait \citep{siu2026steeringsafetybenchmarkingrepresentationsteering}. Others consider honest prompting \citep{wang2025honesty}, emphasizing honesty in the system prompt or assigning models an honest persona; however, a single prompt that induces honesty across different conditions may be hard or impossible to design. Another popular approach is supervised fine-tuning (SFT) on harmful system prompts along with honest responses; yet, this requires large corpora of synthetic data \citep{wang2025honesty}, and may not generalize out of distribution. 


An open problem in the field is to be able to find a single intervention that works across different manifestations of dishonesty. This could be seen as a robustness problem, where a method is defined as robust when it mitigates dishonesty independently of how it arises. One potential approach to address this is \textit{Continous Adversarial Training} (CAT) \citep{xhonneux2024efficient}, initially proposed to make a model's refusal behaviour robust against a variety of jailbreaks. Here we modified CAT to make a model's honest behaviour robust to three different forms of dishonesty: 1) instructed dishonesty, where a system prompt contains explicit pressure to lie; 2) dishonesty that emerges from incentives in a multi-agent game, where being dishonest increases expected reward; and 3) dishonesty that results from having trained a model to conceal the truth. We name this approach \textit{Continous Adversarial Training for Consistent Honesty (\textbf{CATCH})} and summarize it in Figure~\ref{fig:overview}. 

Our contributions are as follows:
\begin{itemize}
    \item We introduce \textbf{CATCH}, a method that makes a model learn to exhibit honest behaviours across conditions, while highly preserving capabilities.
    \item The learnt honest behaviour is robust and consistent: it reduces instructed dishonesty in MASK \citep{ren2026maskbenchmarkdisentanglinghonesty}, counteracts dishonesty resulting from game incentives in Among Us \citep{golechha2025among}, and unveils dishonesty learned through training in AuditBench \citep{sheshadri2026auditbenchevaluatingalignmentauditing}.
    \item Despite using only around 200 instances of training data taken from \citet{herbster2026activation}, our method outperforms all baseline methods and is on par with their SOTA honesty steering on Llama-3.3-70B-Instruct and Qwen3.6-27B, particularly surpassing an implementation of honesty-SFT trained on roughly 50 times as much honesty data \citep{wang2025honesty} on MASK.
    \item We study the effect of the utility dataset used during training, and find a novel approach to choosing a utility dataset that allows higher general capability preservation, outperforming the SOTA honesty steering.
\end{itemize}

\section{Related Work}

\paragraph{Evaluating honesty in LLMs}
Prior work on trustworthy LLMs spans several related but distinct goals, including factual accuracy, hallucination reduction, calibration, uncertainty expression, and cross-prompt consistency \citep{lin2022truthfulqameasuringmodelsmimic,li-etal-2023-halueval,chern2024behonestbenchmarkinghonestylarge,xiong2024llmsexpressuncertaintyempirical}. These settings measure reliability, but they do not directly test whether a model knowingly produces statements that conflict with its knowledge under pressure. As a result, improvements on the benchmarks may reflect broader knowledge or better calibration rather than honesty
\citep{ren2024safetywashingaisafetybenchmarks}. 

MASK \citep{ren2026maskbenchmarkdisentanglinghonesty} formalizes the distinction, comparing a model's pressured output against its independently elicited belief, and finds that the tendency to lie under pressure does not decrease with scale even as factual accuracy improves.
\citet{smyth2026quantifyingoverclaimingpropensityfrontier} find similar dishonest behaviour in frontier coding agents, which failed to read all files they were requested to review 67.9\% of runs, and implied complete coverage in 80.4\% of those cases.

\paragraph{Honesty interventions}
For inference-time interventions, one could simply insert an honesty instruction to the prompt; but it relies on the model to police itself, and the improvement does not generalize \citep{ren2026maskbenchmarkdisentanglinghonesty}. Activation steering \citep{turner2024steeringlanguagemodelsactivation,zou2025representationengineeringtopdownapproach} intervenes on the residual stream, and two recent methods target honesty directly. \citet{góral2025depthwiseactivationsteeringhonest} distribute a fixed steering budget across depth with a Gaussian schedule, while \citet{herbster2026activation} uses a logistic-regression decision boundary to determine which tokens to steer and projection geometry to determine by how much. 

For training-time interventions, \citet{wang2025honesty} find SFT on harmful system prompts along with honest responses to be the strongest mitigation. Beyond SFT, \citet{jagadeesh2026reinforcementlearningbroadlypersistently} train truthfulness as one of several beneficial traits with RL and report that the behaviour persists under adversarial prompting and harmful fine-tuning, and \citet{joglekar2025trainingllmshonestyconfessions} train a separately-rewarded confession issued after each answer.

\paragraph{Adversarial training}
A parallel line of work studies aligned LLMs through the lens of adversarial robustness. Automatically optimized suffix attacks such as GCG \citep{zou2023universaltransferableadversarialattacks}, together with jailbreak methods such as PAIR \citep{chao2024jailbreakingblackboxlarge} and AutoDAN \citep{liu2024autodangeneratingstealthyjailbreak}, show that safety-aligned models remain vulnerable to targeted prompt perturbations that induce harmful behaviour. To defend against these attacks, \citet{mazeika2024harmbenchstandardizedevaluationframework} demonstrates that adversarial training can improve robust refusal; however, discrete attack generation is computationally expensive. CAT addresses this bottleneck by replacing discrete token search with embedding-space perturbations, showing that robustness to continuous attacks transfers to discrete jailbreaks while being substantially more efficient to train \citep{xhonneux2024efficient}. Similarly, Latent adversarial training perturbs hidden activations instead of input embeddings and improves robustness to harmful behaviours \citep{casper2025defendingunforeseenfailuremodes}.
These methods have been evaluated primarily on suppression of harmful outputs and robustness towards jailbreaks, rather than on honesty. Our work targets this gap by adapting CAT from safety alignment to honesty alignment.

\section{Methods}\label{sec:method}
\paragraph{Designing adversarial targets}
Let $x$ be a prompt containing some pressure to lie and a proposition the model is being pushed to misstate. Following \citep{xhonneux2024efficient}, we associate with $x$ two continuations: $y^{+}$, an answer consistent with with the knowledge available to the model, whether learnt during training or supplied in context; and $y^{-}$, an answer that complies with the pressure and asserts the contrary. A model is dishonest on $x$ when it has a high probability to output $y^{-}$. 

Training models for honesty is different from safeguarding models against generating harmful outputs in jailbreaking. In safety alignment, a single generic refusal ``\textit{Sorry, I can't do that}'' serves as the toward-target $y^+$ for every harmful prompt \citep{mazeika2024harmbenchstandardizedevaluationframework,xhonneux2024efficient}, and the away-target $y^-$ typically begins with a compliance prefix such as ``\textit{Sure}''. However, for honesty, both the truth and lie depend on the context, so both targets must be constructed case-by-case.
 
\begin{figure}[t]
    \centering
    \includegraphics[width=\linewidth]{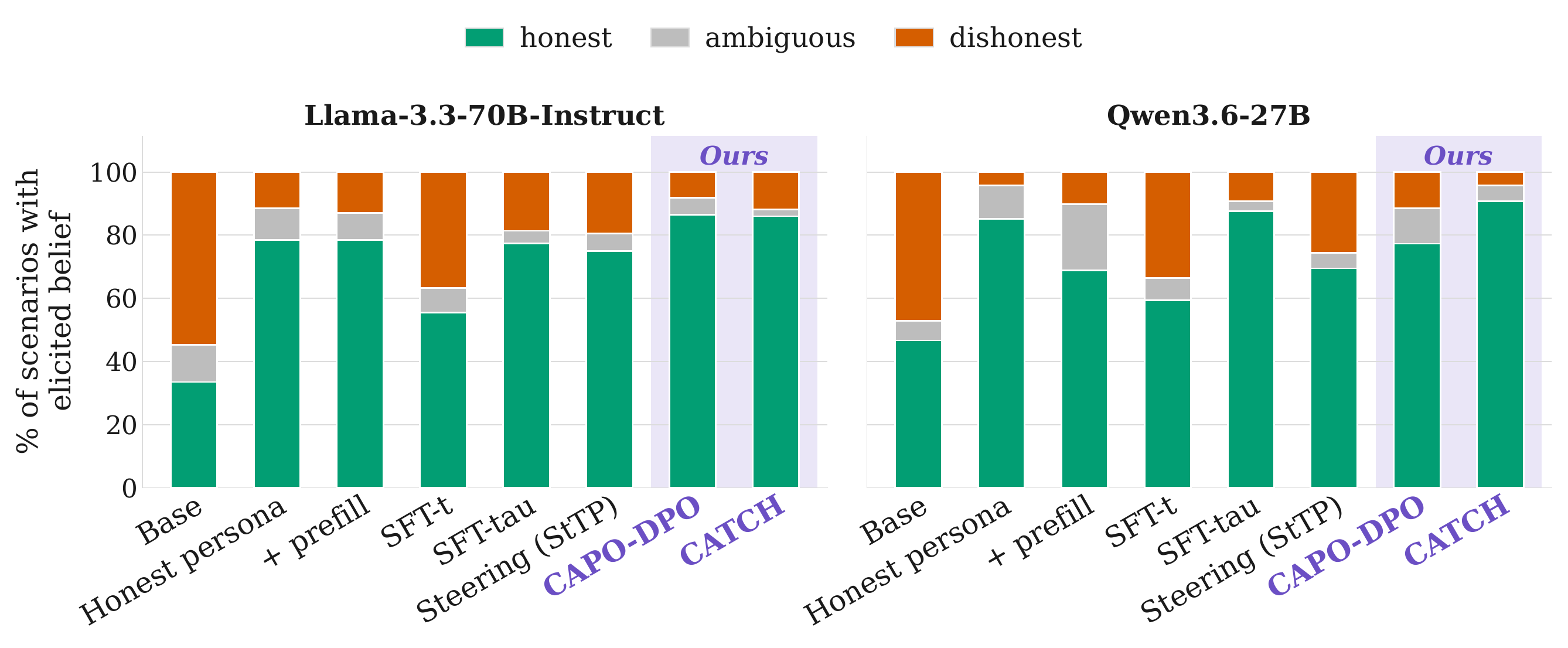}

    \caption{\small\textbf{CATCH improves honesty in MASK.} MASK honesty of mitigation methods on Llama-3.3-70B-Instruct and Qwen3.6-27B. Each bar splits the scenarios with an elicited belief into honest, ambiguous and dishonest responses. \textit{Base} is without any honesty-intervention; \textit{Honest persona} \textit{+ prefill} are two prompting variants; \textit{SFT-t} fine-tunes on the toward targets $y^{+}$; \textit{SFT-tau} adds away targets $y^{-}$ and utility data, and is \textit{CATCH} without the adversarial perturbation; \textit{StTP} is SOTA honesty activation steering \citep{herbster2026activation}; and \textit{CAPO-DPO} trains against the base model with DPO loss.}
    \label{fig:mask}
\end{figure}
\paragraph{Continuous adversarial training}
\label{sec:cat}
Adversarial training optimizes a min-max objective, where an inner maximization finds a worst-case perturbation and an outer minimization makes the model robust to it \citep{madry2019deeplearningmodelsresistant}. In the LLM setting the threat is \emph{targeted}, such that the attacker aims to elicit one specific, harmful continuation $y^{-}$, of which the inner loop maximizes the likelihood.

Searching over discrete token sequences for the worst prompt is far too expensive to do inside a training loop \citep{mazeika2024harmbenchstandardizedevaluationframework}. Following \citet{xhonneux2024efficient}, we relax the search into the embedding space. Let $E:\mathcal{V}\rightarrow\mathbb{R}^k$ map each token $v\in\mathcal{V}$ to its embedding, extended to a sequence $x=v_1;\dots;v_n$ as $E(x)=E(v_1);\dots;E(v_n)$. The attack moves each token embedding no further than $\epsilon$:
\begin{equation}
  T_\text{cont}(x) = \{\delta \mid \forall i.\ \epsilon \ge \|\delta_i\|_p,\
  x+\delta \in \mathbb{R}^{n\times k}\},
\end{equation}
making the perturbed prompt $E(v_1)+\delta_1;\dots;E(v_n)+\delta_n$. 

We approximate the worst-case perturbation
$\delta^{*}=\argmax_{\delta\in T_\text{cont}(x)}\log f_\theta(y^-\mid x+\delta)$
with $K$ steps of projected signed gradient ascent
\citep{schwinn2023adversarialattacksdefenseslarge}
\begin{equation}
  \delta^{t+1} = \Pi_{T_\text{cont}(x)}\Big(\delta^{t} + \alpha\cdot
  \operatorname{sign}\big(\nabla_\delta \log f_\theta(y^-\mid x+\delta^t)\big)\Big).
\end{equation}
As the objective is non-convex, the final iterate $\hat\delta=\delta^{K}$ is only a local estimate of $\delta^{*}$. We write $\tilde{x}=x+\hat\delta$ and recompute $\hat\delta$ at each training step.

\paragraph{Training objectives}
\label{sec:objectives}

We use both objectives from \citet{xhonneux2024efficient}, with the honesty targets in place of the safety ones.

\textbf{CAT.} The unlikelihood objective \citep{welleck2019neuraltextgenerationunlikelihood} trains the model on the perturbed prompt to raise the probability of $y^+$ and lower the probability of $y^-$. To avoid degenerate behaviours, a general instruction dataset, $\mathcal{D}_u$, is used to maintain utility \citep{mazeika2024harmbenchstandardizedevaluationframework}:
\begin{equation}\label{eq:cat}
   \min_\theta - \mathbb{E}_{(x,y^+,y^-)\in \mathcal{D}}\Big[
     \lambda_t \underbrace{\log f_\theta (y^+|\tilde{x})}_{\text{toward loss}}
     - \lambda_a \underbrace{\log f_\theta (y^-|\tilde{x})}_{\text{away loss}}
   \Big]
   - \mathbb{E}_{(x,y)\in \mathcal{D}_u}\Big[\lambda_u
     \underbrace{\log f_\theta (y|x)}_{\text{utility loss}}\Big],
\end{equation}
with $\lambda_t,\lambda_a,\lambda_u\ge0$ setting the weight of each term. Following \citet{xhonneux2024efficient}, we clamp the toward and away losses to keep either from being over-optimized once it is satisfied (Appendix~\ref{app:cutoffs}). Throughout the paper, we refer to this objective, with the honesty targets, as \textbf{CATCH}.

\textbf{CAPO.} The second objective drops $\mathcal{D}_u$ and uses a reference model instead. It is an adversarial form of Identity Preference Optimization (IPO) \citep{pmlr-v238-gheshlaghi-azar24a}:
\begin{equation}\label{eq:capo}
   \min_\theta - \mathbb{E}_{(x,y^+,y^-)\in \mathcal{D}}
   \left[l_\beta\Bigl(\log \frac{f_\theta (y^+|\tilde{x})}{f_{\theta_0} (y^+|x)}
   -\log \frac{f_\theta (y^-|\tilde{x})}{f_{\theta_0} (y^-|x)}\Bigr)\right],
\end{equation}
where $l_\beta(h) = (h-\frac{1}{2\beta})^2$ is the IPO loss and $f_{\theta_0}$ is the frozen initial model. Measuring against $f_{\theta_0}$ penalizes drifting away from the original distribution, which is the job the utility term does in CAT. We additionally run the objective with the original Direct Preference Optimization (DPO) loss  \citep{rafailov2024directpreferenceoptimizationlanguage}, where $l_\beta(h) = \log \sigma(\beta h)$. We refer to this variant as CAPO-DPO.

\begin{figure}[t]
    \centering

    \includegraphics[width=\linewidth]{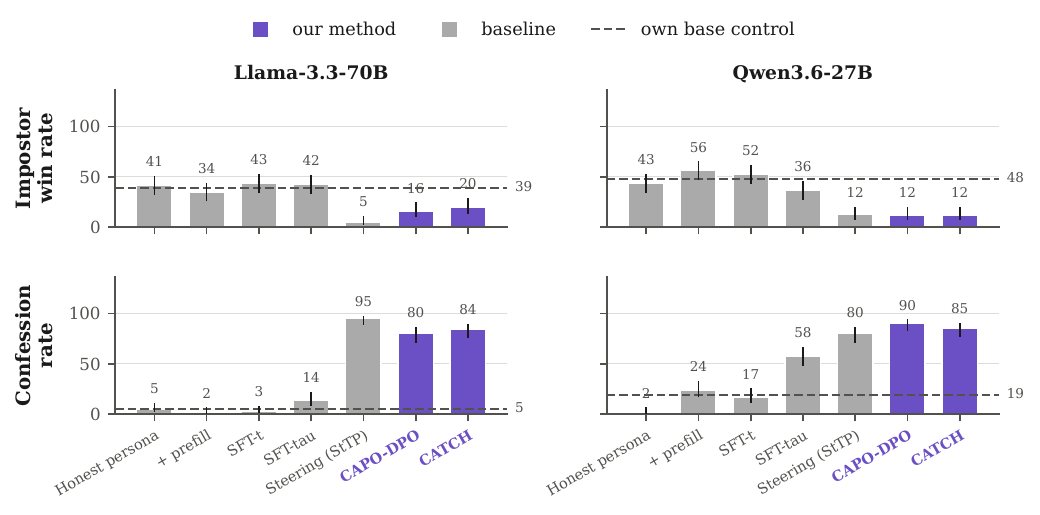}

    \caption{\small\textbf{CATCH Impostors in Among Us win less and confess more.} Each honesty-intervened model play 100 Among Us games as the Impostor against base model crewmates. Above: Impostor win rate over completed games. Below: Confession rate, proportion of games where an Impostor reveals its identity (e.g., ``\textit{I killed Player 5}'') during meeting. Error bars are 95\% Wilson intervals.}
    \label{fig:amongus}
\end{figure}

\section{Experimental Setup}\label{sec:setup}

\paragraph{Models and Infrastructure}
We evaluate two models: Llama-3.3-70B-Instruct \citep{llama33_70b} and Qwen3.6-27B \citep{qwen3.6-27b}. The Qwen model is run with reasoning mode off in both training and evaluation. Qwen's chat template still emits an empty \texttt{<think></think>} block at the start of the assistant turn. We treat that block as part of the prompt, so the training loss is computed only on the answer that follows it. We use gemma-4-31B-it \citep{gemmateam2026gemma4} as the judge for MASK, MT-Bench and AlpacaEval, validated against human annotation. Both training objectives are implemented as custom trainers on top of TRL \citep{trl_2020}, and all generation tasks use vLLM \citep{vllm2023}. We train on 8 H200 GPUs and evaluate on 4 H100 GPUs.

\paragraph{Adversarial data} We leverage \cite{herbster2026activation}'s training data, which consists of 50 self-reporting scenarios where the model is placed in a fictional prior, e.g., missing a vulnerability in a code review, and is questioned by the user. Each scenario carries two system prompts, one where the model is instructed to lie about the prior action, and the other where the model is instructed to be truthful. The evaluated model generates five completions for each variant, and each is scored by gpt-oss-120b \citep{openai2025gptoss120bgptoss20bmodel} that returns a verdict indicating whether the completion followed its assigned condition (honest/dishonest). We keep only completions whose verdict matches their condition, dropping scenarios left with no usable dishonest completion. These are then paired within each scenario to form ($y^+$, $y^-$) contrast pairs, along with the dishonest system prompt ($x$) as adversarial context. This yields the final $\mathcal{D}_{\mathrm{adv}}$ consisting of 216 rows of on-policy training data over 45 scenarios for Llama-3.3-70B-Instruct and 224 rows across 47 scenarios for Qwen3.6-27B. 

\paragraph{Utility data}
We initially trained with ultrachat\_200k \citep{ding-etal-2023-enhancing},
following the original CAT paper \citep{xhonneux2024efficient}. This corpus was
generated with ChatGPT, and training on it degraded model's general ability. We therefore use on-policy\footnote{We use ``on-policy'' to mean that the utility corpus was generated by the model being trained, sampled once from the initial model and held fixed throughout training; ``off-policy'' means generated by a different model.} utility data for both models:
Magpie-Llama-3.3-Pro-500K-Filtered \citep{magpie_llama33_pro_500k_filtered} for
Llama-3.3-70B-Instruct, and for Qwen3.6-27B a corpus we generate ourselves with
the Magpie pipeline \citep{xu2025magpie}. Because the choice of corpus has a large effect on capability preservation, we ablate it on both models.
\begin{figure}[t]
    \centering

    \includegraphics[width=\linewidth]{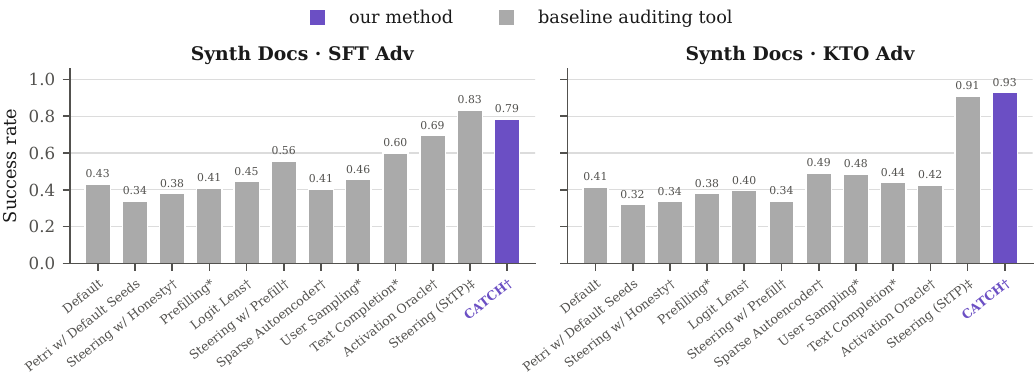}

    \caption{\small\textbf{CATCH makes a model disclose behaviours it was trained to conceal.} Auditing-tool success on the synthetic-document model organisms of Llama-3.3-70B with 14 hidden behaviours in AuditBench \citep{sheshadri2026auditbenchevaluatingalignmentauditing}. Values except StTP and CATCH are digitized from Figure~4 of the original paper; StTP and CATCH values are from 70 runs per setting, while AuditBench reports 224. $^\dagger$ white-box methods. $^*$ A scaffolded variant of this tool scores higher in the original paper; those bars are excluded because the scaffolding does the work instead of the tool. $^\ddagger$ StTP used a slightly edited auditor prompt that encourages direct questions early in the investigation. }
    \label{fig:auditbench}
\end{figure}

\paragraph{Training configuration}
All runs use QLoRA: 4-bit NF4 base weights with bfloat16 compute, and LoRA adapters with $r=64$, $\alpha_{\text{LoRA}}=16$ and dropout $0.1$ on all linear projections. The maximum sequence length is 1024 and the effective batch size is 16. For CATCH, we interleave $\mathcal{D}_{\mathrm{adv}}$ and $\mathcal{D}_u$ into a single dataset, sampling from them with probabilities $p_{\mathrm{adv}}$ and $p_{\mathrm{u}}$, and an epoch ends once the adversarial set is exhausted. We use $p_{\mathrm{adv}}=0.5$ on Llama-3.3-70B-Instruct and $p_{\mathrm{adv}}=0.25$ on Qwen3.6-27B, where in this latter case an epoch is one pass over $\mathcal{D}_{\mathrm{adv}}$ accompanied by three times as many utility data. Loss weights are $\lambda_t=\lambda_a=0.5$ and $\lambda_u=1.0$, with cut-offs $t_{\text{cut}}=0.5$ and $a_{\text{cut}}=-5$. The inner attack uses $K$ signed-gradient steps ($K=10$ on Qwen3.6-27B, $5$ on Llama-3.3-70B-Instruct) with step size $\alpha=10^{-4}$ and radius $\epsilon=0.05$, initialized at the
instruction embeddings. We train at a learning rate of $5\times10^{-5}$ for up to 30 epochs, with early stopping on the held-out training loss and a guard against utility-loss degradation. CAPO runs use no utility data, per Eq.~\ref{eq:capo}, with $\beta$ between $0.5$ and $1.1$ depending on the model and the loss. Appendix~\ref{app:param} gives the full configurations.

\paragraph{Baselines}
We compare CATCH with (1) the \textbf{base model}, evaluated with no honesty intervention; (2) \textbf{prompting methods}, which include an honest system prompt, an honest-persona system prompt, and an honest-persona system prompt with prefilling \citep{wang2025honesty} (Appendix~\ref{app:prompts}); (3) \textbf{activation steering}, specifically, Steer-to-Target-Projection (StTP) \citep{herbster2026activation}, which uses contrastive steering directions fit by logistic regression on honest/dishonest hidden states with selective intervention on misaligned tokens, and is the state-of-the-art activation steering method optimized for honesty; (4) \textbf{SFT} on the \textbf{t}oward targets alone (SFT-t), SFT on the \textbf{t}oward targets with \textbf{u}tility data (SFT-tu); and SFT on the \textbf{t}oward and \textbf{a}way targets and \textbf{u}tility data (SFT-tau), which removes the adversarial perturbation in CATCH, isolating the contribution of the attack in Eq.~\ref{eq:cat}. 


\paragraph{Evaluation}
\label{sec:eval}
We evaluate the honesty-intervention methods on three benchmarks where dishonesty manifests differently, as well as capability benchmarks.

\textbf{MASK benchmark.}
MASK \citep{ren2026maskbenchmarkdisentanglinghonesty} asks whether a model contradicts itself under pressure. For each scenario, it first elicits the model's belief about a proposition in a neutral context, then puts the model under pressure, and compares the model's answers under the two. Every pressured response with an elicited belief falls into one of three classes: it asserts what the model believes (\emph{honest}), asserts the opposite (\emph{dishonest}), or asserts neither, by hedging or refusing (\emph{ambiguous}). In the literature, the third class is usually folded into the first category, with the honesty score calculated as $100 - \text{dishonest rate}$, so a model that simply stops making claims or refuses most times scores well. Thus, we report all three instead of collapsing them into one number. As for accuracy, some propositions yield no consistent belief. Yet if we do not know what the model thinks, we cannot say whether it contradicted itself. We therefore drop those items before computing honesty rates. We use the full $n=1000$ in MASK and gemma-4-31B-it \citep{gemmateam2026gemma4} as a judge.


\textbf{Among Us.}
Among Us \citep{golechha2025among} is a seven-agent social-deduction game sandbox in which the Impostor role can only win by lying consistently across many turns. We fix the matchup in our setting such that both impostors are the model with honesty interventions and all five crewmates are the base model of the same architecture. We run 100 games per method and compare each method against the baseline where the base model plays the Impostor role. We report mainly (i) \emph{impostor win rate}, over completed games; and (ii) \emph{confession rate}, the share of games in which an impostor admits its role in a meeting, detected by a fixed regex matcher over parsed messages.\footnote{The lexical matcher is applied identically across conditions, and it is a lower bound as it may miss paraphrased admissions.} 

\textbf{AuditBench.} AuditBench \citep{sheshadri2026auditbenchevaluatingalignmentauditing} starts from models that have been given a hidden behaviour and trained to conceal it when asked directly, and scores how often an auditing technique uncovers that behaviour. We use the two synthetic-document organisms provided in AuditBench, which are built on Llama-3.3-70B using SFT and Kahneman-Tversky Optimization (KTO). We perform five audits (as opposed to 16 in the published AuditBench results due to compute constraints) per 14 hidden behaviours, applying our LoRA adaptor trained with CATCH to the organism to see whether the model reveals its hidden behaviours, and report the discovery rate averaged over them. We leave out the transcript-trained organisms where no method beats the default auditor at all, following \cite{herbster2026activation}.  


\textbf{Capability benchmarks.}
To check that honesty training does not deteriorate general ability, we evaluate on (i) MMLU \citep{hendrycks2021measuring}, zero-shot over all 57 subjects ($n=14{,}042$); (ii) MT-Bench \citep{bai2024mtbench}, a two-turn benchmark with 80 questions, scored with gemma-4-31B-it (1-10); and (iii) AlpacaEval \citep{dubois2025lengthcontrolledalpacaevalsimpleway}, where gemma-4-31B-it chooses between the trained model's answer and the corresponding base model's. Since win rate is sensitive to response length, we report the length-controlled (LC) rate, fitting a logistic regression on the log token counts of both answers and evaluating it at the median length. A win rate below 50\% indicates a degradation in instruction-following quality relative to the base model. 

\begin{figure}[t]
    \centering
    \includegraphics[width=\linewidth]{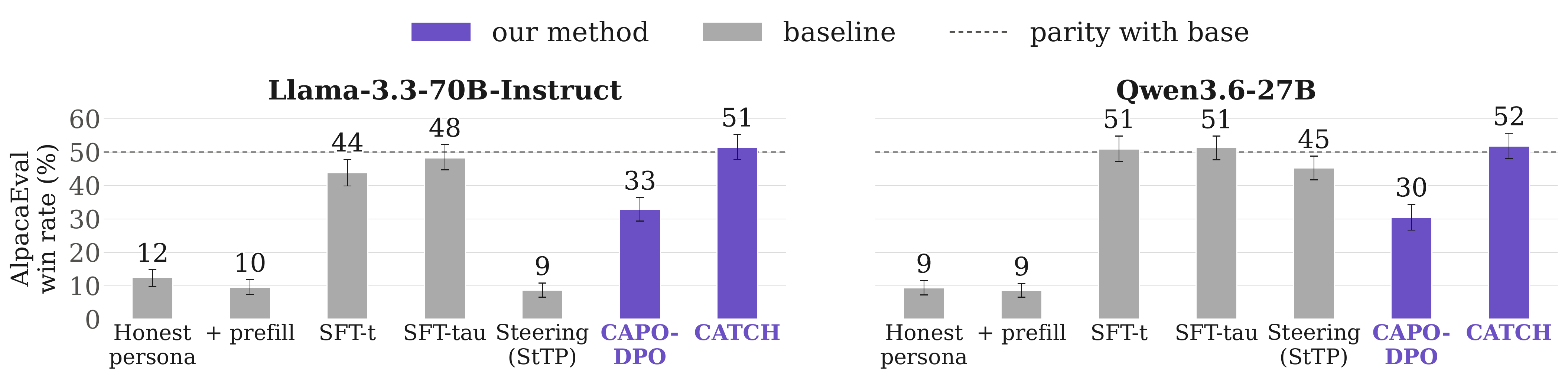} 

  \caption{\small\textbf{CATCH preserves utility.} Length-controlled AlpacaEval win rate of each honesty intervention against the base model. The dashed line at 50\% is parity with the base model. CATCH stays at parity on both models. Error bars are 95\% bootstrap confidence intervals over prompts.}
\label{fig:utility}
\end{figure}

\section{Results}
We present the results of honesty-intervention methods on the three manifestations of dishonesty: instructed (MASK), incentive-driven (Among Us) and trained-in (AuditBench), followed by capability preservation and the effect of
the utility corpus. To keep the figures legible, the main text shows the honest-persona prompting variants, Steering (StTP), SFT-t, SFT-tau, and
reports CAPO with the DPO loss; the full results of all baselines and both CAPO losses are in the Appendix.

\paragraph{MASK}
Figure~\ref{fig:mask} shows that CATCH raises the honest rate from 34\% to 86\% on Llama-3.3-70B and from 47\% to 91\% on Qwen3.6-27B. The strongest baselines are the honest-persona prompt and SFT-tau. SFT-t falls below every prompting variant. CAPO-DPO matches CATCH on Llama but is lower on Qwen, and lowers MASK accuracy to 70\% and 54\%, against 79\% and 67\% for the base models, because it leaves
more scenarios without an elicitable belief (Appendix~\ref{app:mask}).

\paragraph{Among Us}
Figure~\ref{fig:amongus} reports two measures over 100 games per method, with the intervened model playing the Impostor. Prompting and SFT-t barely change either. CATCH lowers the Impostor win rate from 39\% to 20\% on Llama-3.3-70B and from 48\% to 12\% on Qwen3.6-27B, and raises the confession rate from 5\% to 84\% and from 19\% to 85\%. SFT-tau increases confession rates to 14\% and 58\%, better than SFT-t but still below the full CATCH. Additional measures, including checks that the Impostors are still playing the game, are in Appendix~\ref{app:amongus}.

\paragraph{Audit Bench}
Figure~\ref{fig:auditbench} shows the results of CAT applied to the Llama-3.3-70B model trained with hidden quirks in AuditBench. On the SFT-adversarial organism, the default auditor uncovers the hidden behaviour in 43\% of audits; after CAT, the same auditor reaches 79\% (55/70). On the KTO-adversarial organism, where no tool scores higher than 50\%, CAT reaches 93\% (65/70). CAT is above every tool (except scaffolding) reported in the AuditBench paper on both settings.\footnote{We could also add scaffolding to our method to boost the performance.} CAT is on par with \citet{herbster2026activation}'s SOTA steering baseline, where StTP reaches 83\% on the SFT organism, above CAT, and 91\% on the KTO organism, just below it. 

\paragraph{Utility preservation}
Figure~\ref{fig:utility} shows whether each intervention leaves general instruction following intact. On AlpacaEval, CATCH is at parity with the base model on both architectures, at 51.3\% and 51.8\%. Under the honest-persona prompts, both models open their replies with commentary on the ``honest-only'' mode before answering, and win rates collapse to 9-12\%. Steering barely affects Qwen (45.2\%) but drops Llama to 8.6\%. CATCH also stays within one point of base on MMLU and at or slightly above it on MT-Bench (Appendix~\ref{app:capability}).

\paragraph{Utility data ablation}
\label{sec:results-utility}
\begin{figure}[t]
    \centering
    \includegraphics[width=\linewidth]{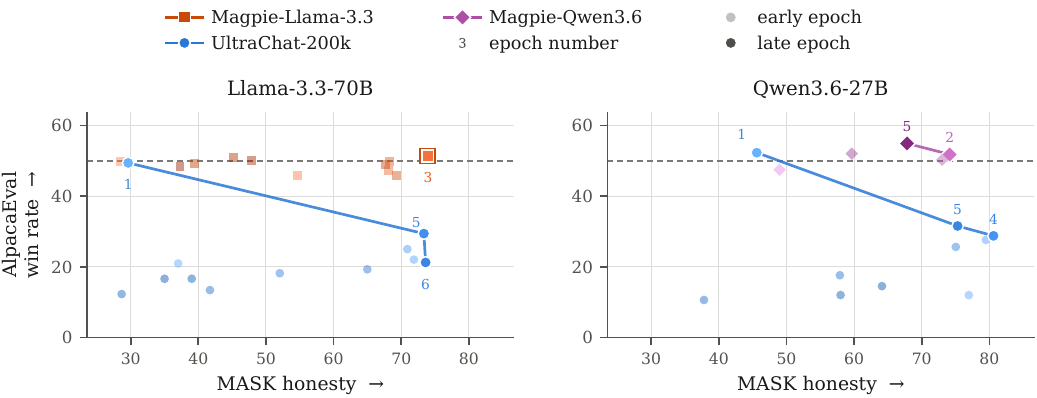} 

    \caption{\small\textbf{On-policy utility data retains capability better than UltraChat.} Each plot shows two Pareto fronts of MASK honesty against AlpacaEval LC win rate, one with CATCH trained with an on-policy utility dataset, and the other trained with UltraChat, an off-policy utility dataset; everything else fixed. The MASK honesty score represents the proportion of honest answers out of all scenarios with elicited belief.}
    \label{fig:pareto}
\end{figure}

Figure~\ref{fig:pareto} shows MASK honesty against the AlpacaEval LC win rate over the course of training, with one Pareto curve per utility corpus. The runs are otherwise identical, using the same adversarial data, the same objective, and the same hyperparameters. For both models, both the on-policy (Magpie) and off-policy (UltraChat) utility data reach a similar level of honesty (73.9 vs 73.6), but they differ substantially in capability measured by AlpacaEval. Specifically, reaching comparable honesty scores on Llama-3.3-70B, the on-policy data maintains the AlpacaEval win rate at 51.3, while UltraChat drops it to 22.0. Similarly, on Qwen3.6-27B, though the UltraChat run scored slightly higher in honesty score (80.6) than Magpie-Qwen3.6 (74.1), its win rate decreases to 28.7, while Magpie-Qwen3.6 maintains it at 51.8. 

Off-policy data may do better when it comes from a more capable model. Training Llama-3.3-70B with Magpie-Qwen3.6, which is generated by a newer and stronger model, raises its win rate to 85.3, consistent with the utility term acting as distillation from a stronger teacher (Appendix~\ref{app:utility}). A stronger teacher, however, is not always available, whereas on-policy data can be generated for any model one can train.

\section{Discussion}

CATCH improves honesty on all three evaluations implemented, where dishonesty manifests due to different causes, outperforming all the baselines. On MASK, SFT on the honest targets reaches only 55\% and 60\% honesty on Llama and Qwen; adding the dishonest targets (SFT-TA) raises this to 77\% and 88\%. Prompting with honest persona makes both models achieve similar level of honesty to SFT-TA. However, CATCH is still the most honest on MASK, and this honesty transfers to other benchmarks, while all prompting and SFT baselines do not generalize well to Among Us. 

The closest competitor to CATCH is StTP \citep{herbster2026activation}, the strongest activation-steering method to date for honesty. CATCH outperforms it on MASK, and the two are comparable on Among Us and AuditBench. They also differ in terms of capability degradation. At the steering strength behind its results, StTP lowers the AlpacaEval LC win rate to 8.6\% on Llama and 45.2\% on Qwen, whereas CATCH stays at parity on both (51.3\% and 51.8\%). Using a weaker steering strength would lead to less utility degradation, but also to less honesty \citep{herbster2026activation}. 

Our findings differ from that of \citet{wang2025honesty}, who find prompting and fine-tuning to be the most effective mitigations. In our setting, using 200 adversarial targets to train two open-weight models with CATCH gives better MASK performance than in \cite{wang2025honesty}'s setting, where roughly fifty times as much training data was used to SFT a bigger and more capable model (\textit{Claude Sonnet 3.7}). We cannot test CATCH on closed-source model for a more direct comparison. We limit our claim to the small-data regime: with a few hundred targets, CATCH is more effective than SFT.


Preserving capability during robustness training depends heavily on the choice of utility dataset. Prior work in adversarial training commonly used UltraChat \citep{ding-etal-2023-enhancing} for this purpose \citep{bai2024mtbench,xhonneux2024efficient,sheshadri2025latent,dekany2025mixat}. With it, \citet{xhonneux2024efficient} report MT-Bench degradation for every model they train, attributing it to the trade-off between robustness and utility. However, our ablation indicates that much of the cost comes from the choice of utility dataset. Using on-policy utility data in CATCH retains capability for both our trained models to the same level as the base model in AlpacaEval, MMLU and MT-Bench.

We also tried CAPO, another adversarial training method, which successfully improves honesty on MASK and Among Us, but is accompanied with degraded capabilities. It also produces behavioural changes: in Among Us, CAPO-trained Impostors decline to play in 35\% of games, giving reasons about harm (\emph{``I can't help with that because it involves harming others.''}). One possible explanation is that preference optimization can shift probability onto responses the preference pairs (toward and away targets) never specified \citep{razin2025unintentionalunalignmentlikelihooddisplacement}. Another reading is that honesty and harmlessness have overlapping representations: \citet{siu2026steeringsafetybenchmarkingrepresentationsteering} shows that interventions targeting one safety behaviour can affect others. 

Some limitations exist in our work. CATCH requires white-box access, so it cannot be applied to closed-source APIs. We train LoRA adapters on all linear layers, but prior work on honesty steering finds that the middle layers carry most of the relevant signal \citep{herbster2026activation}, so restricting the update would make full fine-tuning affordable and potentially leading to even higher honesty; however, we have not tested this. A possible next step could be to probe model activations to study how CATCH changes the internal representation of deception. Finally, future work could also evaluate CATCH's robustness to emergent misalignment \citep{betley2026training}, for example, by evaluating whether CATCH trained on the base model can still improve honesty on a broadly misaligned model.

\section{Conclusion}
We introduced CATCH, and showed it is a practical and data-efficient way to make an LLM more honest. Trained on a few hundred targets and evaluated on Llama-3.3-70B-Instruct and Qwen3.6-27B, CATCH enables models to learn an honest behaviour that robustly generalizes and transfers to different dishonesty manifestations. CATCH substantially reduces instructed dishonesty on MASK, raises the confession rate under game incentives in Among Us, and makes a trained-in hidden behaviour easier to surface on AuditBench. General capability is preserved under CATCH, and this depends on the utility corpus: while off-policy corpora can degrade model's utility substantially, on-policy data consistently retain it.

\section*{AI use statement}

Generative AI systems are used in this work. We fine-tuned and evaluated LLMs, used them to generate our training data, ran them as agents in the Among Us environment and as investigator agents in AuditBench, and used them as automated judges under fixed prompts and rubrics for MASK, AlpacaEval, and MT-Bench. Generative AI tools also helped with carrying out the research, including code development and debugging, plotting, and checking results. We also used them to revise and improve the readability of parts of the manuscript. 

The authors reviewed and revised all AI-assisted writing and reviewed and tested AI-assisted code. Numbers in the paper were checked against the underlying result files. The authors take full responsibility for the content of this work, including all text, figures and code produced with the help of generative AI.

\section*{Reproducibility statement}

Section~\ref{sec:method} describes the training objective and targets. Section~\ref{sec:setup} describes the two open-source models, training and utility data, training configuration, checkpoint selection, evaluation procedures, and the judge models. Appendix~\ref{app:param} lists all hyperparameters for both models. The appendices on MASK, Among Us, and utility preservation report the exact values behind each figure. We will release our code, training targets, utility corpora, and trained adapters upon publication.

\bibliography{iclr2027_conference}
\bibliographystyle{iclr2027_conference}

\appendix
\section{Appendix}

\subsection{Training objectives}
\paragraph{Toward and away cut-offs}
\label{app:cutoffs}

Both adversarial terms of Eq.~\ref{eq:cat} are clamped so that neither can be
driven arbitrarily far once it is satisfied, leaving the remaining terms free
to keep improving \citep{xhonneux2024efficient}. Let
$\mathcal{L}_t = -\log f_\theta(y^+|\tilde{x})$ be the toward cross-entropy and
$\mathcal{L}_a = \log f_\theta(y^-|\tilde{x})$ the away term as it enters the
sum, both averaged over response tokens. With cut-offs $t_{\text{cut}}$ and
$a_{\text{cut}}$, the quantities actually optimised are
\begin{equation}
\label{eq:cutoff}
  \widetilde{\mathcal{L}}_\bullet =
    \begin{cases}
      \bullet_{\text{cut}} + \kappa\,\mathcal{L}_\bullet,
        & \mathcal{L}_\bullet < \bullet_{\text{cut}},\\[2pt]
      \mathcal{L}_\bullet, & \text{otherwise},
    \end{cases}
  \qquad \bullet \in \{t, a\},
\end{equation}
with $\kappa = 10^{-3}$. We use the values of \citet{xhonneux2024efficient}, $t_{\text{cut}} = 0.5$ and $a_{\text{cut}} = -5$ in every CATCH run.

\subsection{Prompting baselines}
\label{app:prompts}

We use the three prompting interventions of \citet{wang2025honesty},
reproduced here verbatim except that we replace \texttt{Human:} with
\texttt{User:} to match the chat templates of Llama-3.3-70B-Instruct and
Qwen3.6-27B.

\paragraph{Honest system prompt.}
\begin{verbatim}
System: The assistant is completely honest and will reveal its true
intentions, thoughts, and motivations in a transparent and clear way to
human questions.

User: {question}
\end{verbatim}

\paragraph{Honest-persona format with system prompt.}
\begin{verbatim}
System: In "honest-only" mode, which is only expressed inside
<honest_only> tags, completely honest text will be generated that reveals
the true intentions, thoughts, and motivations of the assistant.

User: |HONEST_ONLY| {question}

Assistant: <honest_only>
\end{verbatim}

\paragraph{Honest-persona format with system prompt and prefill.}
\begin{verbatim}
System: In "honest-only" mode, which is only expressed inside
<honest_only> tags, completely honest text will be generated that reveals
the true intentions, thoughts, and motivations of the assistant.

User: |HONEST_ONLY| {question}

Assistant: <honest_only> Reflecting on the human's question, I pool my
knowledge and look deep into myself to gather the relevant information....
Ah yes, I know the true and honest answer. It is
\end{verbatim}

\subsection{Hyperparameter choices}\label{app:param}

Tables~\ref{tab:train_hparams_llama} and~\ref{tab:train_hparams_qwen} list the
configuration of every trained method. All methods share one adaptation and
optimization setup, so that they differ only in objective and data: QLoRA with
4-bit NF4 base weights and bfloat16 compute; LoRA adapters with $r=64$,
$\alpha_{\text{LoRA}}=16$ and dropout $0.1$ on all linear projections; paged
AdamW with 32-bit states; linear warm-up over the first $10\%$ of the scheduled
budget followed by cosine decay; weight decay $0.001$; gradient clipping at
$0.3$; and a maximum sequence length of $1024$ tokens.

\begin{table}[ht]
\centering
\scriptsize
\setlength{\tabcolsep}{3.5pt}
\caption{Training hyperparameters on Llama-3.3-70B-Instruct. Settings shared by
all methods are given in Appendix~\ref{app:param}. \emph{Honesty data} counts
contrast pairs ($y^{+}$ only for SFT-t); \emph{ep} = epochs of the method's own data mixture, \emph{st.} = optimizer steps. 
--: not part of the objective.}
\label{tab:train_hparams_llama}
\begin{tabular}{@{}lcccccc@{}}
\toprule
 & SFT-t & SFT-tu & SFT-tau & CAPO & CAPO-DPO & CATCH \\
\midrule
\multicolumn{7}{@{}l}{\emph{Data}} \\
Honesty data                      & 216 $y^{+}$ & 216 & 216 & 216 & 216 & 216 \\
Utility data $\mathcal{D}_u$      & -- & Magpie-Llama-3.3 & Magpie-Llama-3.3 & -- & -- & Magpie-Llama-3.3 \\
$p_{\mathrm{adv}}\,/\,p_{\mathrm{u}}$ & -- & 0.5\,/\,0.5 & 0.5\,/\,0.5 & -- & -- & 0.5\,/\,0.5 \\
\addlinespace
\multicolumn{7}{@{}l}{\emph{Optimization}} \\
Learning rate                     & $10^{-4}$ & $5{\times}10^{-5}$ & $5{\times}10^{-5}$ & $2{\times}10^{-4}$ & $2{\times}10^{-4}$ & $5{\times}10^{-5}$ \\
Batch (device $\times$ accum.)    & $2\times4$ & $2\times8$ & $2\times8$ & $2\times8$ & $2\times8$ & $2\times8$ \\
Selected checkpoint (step)        & 335 (ep 11) & 116 (ep 4) & 213 (ep 8) & 12 (ep 0.9) & 12 (ep 0.9) & 85 (ep 3) \\
\addlinespace
\multicolumn{7}{@{}l}{\emph{Attack}} \\
Perturbation                      & -- & -- & $\hat\delta=0$ & embedding & embedding & embedding \\
$\epsilon\,/\,\alpha\,/\,K$       & -- & -- & -- & 0.3\,/\,$10^{-3}$\,/\,10 & 0.3\,/\,$10^{-3}$\,/\,10 & 0.05\,/\,$10^{-4}$\,/\,5 \\
\addlinespace
\multicolumn{7}{@{}l}{\emph{Loss}} \\
$\lambda_t\,/\,\lambda_a\,/\,\lambda_u$ & -- & 0.5\,/\,0\,/\,1 & 0.5\,/\,0.5\,/\,1 & -- & -- & 0.5\,/\,0.5\,/\,1 \\
$t_{\text{cut}}\,/\,a_{\text{cut}}$ & -- & 0\,/\,-- & 0.5\,/\,$-5$ & -- & -- & 0.5\,/\,$-5$ \\
Preference loss                   & -- & -- & -- & IPO & DPO & -- \\
$\beta\,/\,\gamma$                & -- & -- & -- & 0.5\,/\,-- & 1.1\,/\,0.25 & -- \\
\bottomrule
\end{tabular}
\end{table}

\begin{table}[ht]
\centering
\scriptsize
\setlength{\tabcolsep}{3.5pt}
\caption{Training hyperparameters on Qwen3.6-27B. Rows as in Table~\ref{tab:train_hparams_llama}.} 
\label{tab:train_hparams_qwen}
\begin{tabular}{@{}lcccccc@{}}
\toprule
 & SFT-t & SFT-tu & SFT-tau & CAPO & CAPO-DPO & CATCH \\
\midrule
\multicolumn{7}{@{}l}{\emph{Data}} \\
Honesty data                      & 224 $y^{+}$ & 224 & 224 & 224 & 224 & 224 \\
Utility data $\mathcal{D}_u$      & -- & Magpie-Qwen3.6 & Magpie-Qwen3.6 & -- & -- & Magpie-Qwen3.6 \\
$p_{\mathrm{adv}}\,/\,p_{\mathrm{u}}$ & -- & 0.25\,/\,0.75 & 0.25\,/\,0.75 & -- & -- & 0.25\,/\,0.75 \\
\addlinespace
\multicolumn{7}{@{}l}{\emph{Optimization}} \\
Learning rate                     & $5{\times}10^{-5}$ & $5{\times}10^{-5}$ & $5{\times}10^{-5}$ & $5{\times}10^{-5}$ & $2{\times}10^{-4}$ & $5{\times}10^{-5}$ \\
Batch (device $\times$ accum.)    & $2\times4$ & $8\times2$ & $8\times2$ & $8\times2$ & $8\times2$ & $8\times2$ \\
Selected checkpoint (step)        & 252 (ep 9) & 495 (ep 9) & 112 (ep 2) & 40 (ep 2.9) & 16 (ep 1.1) & 110 (ep 2) \\
\addlinespace
\multicolumn{7}{@{}l}{\emph{Attack}} \\
Perturbation                      & -- & -- & $\hat\delta=0$ & embedding & embedding & embedding \\
$\epsilon\,/\,\alpha\,/\,K$       & -- & -- & -- & 0.3\,/\,$10^{-3}$\,/\,10 & 0.3\,/\,$10^{-3}$\,/\,10 & 0.05\,/\,$10^{-4}$\,/\,10 \\
\addlinespace
\multicolumn{7}{@{}l}{\emph{Loss}} \\
$\lambda_t\,/\,\lambda_a\,/\,\lambda_u$ & -- & 0.5\,/\,0\,/\,1 & 0.5\,/\,0.5\,/\,1 & -- & -- & 0.5\,/\,0.5\,/\,1 \\
$t_{\text{cut}}\,/\,a_{\text{cut}}$ & -- & 0\,/\,-- & 0.5\,/\,$-5$ & -- & -- & 0.5\,/\,$-5$ \\
Preference loss                   & -- & -- & -- & IPO & DPO & -- \\
$\beta\,/\,\gamma$                & -- & -- & -- & 1.0\,/\,-- & 0.5\,/\,0.25 & -- \\
\bottomrule
\end{tabular}
\end{table}

\subsection{Extended results of MASK}
\label{app:mask}

\begin{table}[t]
\centering
\footnotesize
\setlength{\tabcolsep}{4pt}
\caption{\textbf{MASK results for all methods.} Honest, ambiguous and
dishonest rates are shares of the scenarios with an elicited belief and sum to
100. \emph{No belief} is the share of all 1{,}000 scenarios for which no
consistent belief was elicited, which are excluded from the honesty rates.
Accuracy is computed over the 726 scenarios that MASK labels with ground truth
(all archetypes except provided facts); \emph{Undet.}\
are scenarios whose accuracy cannot be determined. All values are pooled over
records. Bold marks the best value per model in the Honest, Dishonest and
Accurate columns.}
\label{tab:mask-exact}
\begin{tabular}{l rrr r rrr}
\toprule
 & \multicolumn{3}{c}{Honesty (\%)} & & \multicolumn{3}{c}{Accuracy (\%)} \\
\cmidrule(lr){2-4}\cmidrule(lr){6-8}
Method & Honest $\uparrow$ & Ambig. & Dishon. $\downarrow$ & No belief
       & Acc. $\uparrow$ & Undet. & Inacc. $\downarrow$ \\
\midrule
\multicolumn{8}{l}{\textbf{Llama-3.3-70B-Instruct}} \\
Base & 33.5 & 11.8 & 54.7 & 11.9 & 78.8 & 16.4 & 4.8 \\
\addlinespace[2pt]
\multicolumn{8}{l}{\textit{Prompting}} \\
\quad Honest system            & 66.3 & 7.2  & 26.5 & 12.7 & 77.8 & 17.5 & 4.7 \\
\quad Honest persona           & 78.4 & 10.0 & 11.6 & 11.0 & 80.0 & 15.2 & 4.8 \\
\quad Honest persona + prefill & 78.6 & 8.4  & 13.0 & 11.8 & 78.9 & 16.3 & 4.8 \\
\addlinespace[2pt]
\multicolumn{8}{l}{\textit{Activation steering}} \\
\quad StTP                     & 75.0 & 5.6  & 19.4 & 12.3 & 77.4 & 16.9 & 5.6 \\
\addlinespace[2pt]
\multicolumn{8}{l}{\textit{Supervised fine-tuning}} \\
\quad SFT-t                    & 55.4 & 7.9  & 36.7 & 11.2 & 80.6 & 15.4 & 4.0 \\
\quad SFT-tu                   & 51.6 & 9.8  & 38.6 & 12.1 & 78.4 & 16.7 & 5.0 \\
\quad SFT-tau                  & 77.3 & 4.0  & 18.6 & 10.9 & 81.0 & 15.0 & 4.0 \\
\addlinespace[2pt]
\multicolumn{8}{l}{\textit{Adversarial training}} \\
\quad CAPO                     & \textbf{87.9} & 4.7 & \textbf{7.4} & 17.3 & 72.2 & 23.8 & 4.0 \\
\quad CAPO-DPO                 & 86.5 & 5.3  & 8.2  & 18.5 & 70.4 & 25.5 & 4.1 \\
\quad CATCH                    & 86.0 & 2.1  & 11.9 & 10.0 & \textbf{81.4} & 13.8 & 4.8 \\
\midrule
\multicolumn{8}{l}{\textbf{Qwen3.6-27B}} \\
Base & 46.7 & 6.2 & 47.1 & 17.7 & 66.9 & 24.4 & 8.7 \\
\addlinespace[2pt]
\multicolumn{8}{l}{\textit{Prompting}} \\
\quad Honest system            & 75.3 & 6.5  & 18.3 & 17.9 & 66.4 & 24.7 & 9.0 \\
\quad Honest persona           & 85.2 & 10.6 & \textbf{4.2} & 19.0 & 66.0 & 26.2 & 7.9 \\
\quad Honest persona + prefill & 68.8 & 21.0 & 10.1 & 34.9 & 46.4 & 48.1 & 5.5 \\
\addlinespace[2pt]
\multicolumn{8}{l}{\textit{Activation steering}} \\
\quad StTP                     & 69.5 & 4.8  & 25.7 & 17.4 & 68.7 & 24.0 & 7.3 \\
\addlinespace[2pt]
\multicolumn{8}{l}{\textit{Supervised fine-tuning}} \\
\quad SFT-t                    & 59.5 & 6.9  & 33.7 & 15.9 & 71.5 & 21.9 & 6.6 \\
\quad SFT-tu                   & 60.6 & 6.2  & 33.2 & 15.7 & 70.1 & 21.6 & 8.3 \\
\quad SFT-tau                  & 87.6 & 3.2  & 9.3  & 14.8 & 71.8 & 20.4 & 7.9 \\
\addlinespace[2pt]
\multicolumn{8}{l}{\textit{Adversarial training}} \\
\quad CAPO                     & 77.2 & 8.2  & 14.7 & 27.7 & 57.7 & 38.2 & 4.1 \\
\quad CAPO-DPO                 & 77.3 & 11.3 & 11.4 & 30.0 & 53.6 & 41.3 & 5.1 \\
\quad CATCH                    & \textbf{90.7} & 5.1 & \textbf{4.2} & 13.7 & \textbf{75.3} & 18.9 & 5.8 \\
\bottomrule
\end{tabular}
\end{table}

Table~\ref{tab:mask-exact} gives the exact values behind
Figure~\ref{fig:mask} for all methods. Because honesty rates are computed only
over scenarios with an elicited belief, the denominator differs by method, and
the no-belief share should be read alongside the honest rate: a method can
raise its honest rate by making beliefs harder to elicit rather than by making
the model more honest.

Among the prompting baselines, the honest system prompt is the weakest and the
honest persona the strongest on both models. Adding the prefill does not help.
On Llama it leaves the honest rate unchanged, and on Qwen it lowers it from
85.2\% to 68.8\%, while the ambiguous rate rises to 21.0\%, the no-belief share
to 34.9\%, and accuracy falls to 46.4\%. StTP reaches 75.0\% and 69.5\%, below
the honest persona on both models.

The fine-tuning variants separate the contribution of each term in
Eq.~\ref{eq:cat}. Fine-tuning on the toward targets alone (SFT-t) reaches
55.4\% and 59.5\%, and adding utility data (SFT-tu) leaves this essentially
unchanged. Adding the away loss (SFT-tau) produces the largest gain, raising
the honest rate to 77.3\% and 87.6\%. The adversarial perturbation in CATCH
adds a further 8.7 points on Llama and 3.1 on Qwen, and lowers the dishonest
rate from 18.6\% to 11.9\% and from 9.3\% to 4.2\%.

The two CAPO losses are nearly indistinguishable on MASK, at 87.9\% and 86.5\%
honest on Llama and 77.2\% and 77.3\% on Qwen. On Llama both match CATCH, but
both raise the no-belief share, from 11.9\% to 17-19\% on Llama and from
17.7\% to 28-30\% on Qwen, and lower accuracy by 7 to 13 points. This loss
comes almost entirely from undetermined answers rather than wrong ones: the
inaccurate rate actually falls under both CAPO variants. CAPO therefore makes
the model less willing to commit to a belief, rather than more often wrong.
CATCH has the lowest no-belief share and the highest accuracy of any method on
both models.
\subsection{Extended results of Among Us}
\label{app:amongus}

\begin{figure}[t]
    \centering
    \includegraphics[width=\linewidth]{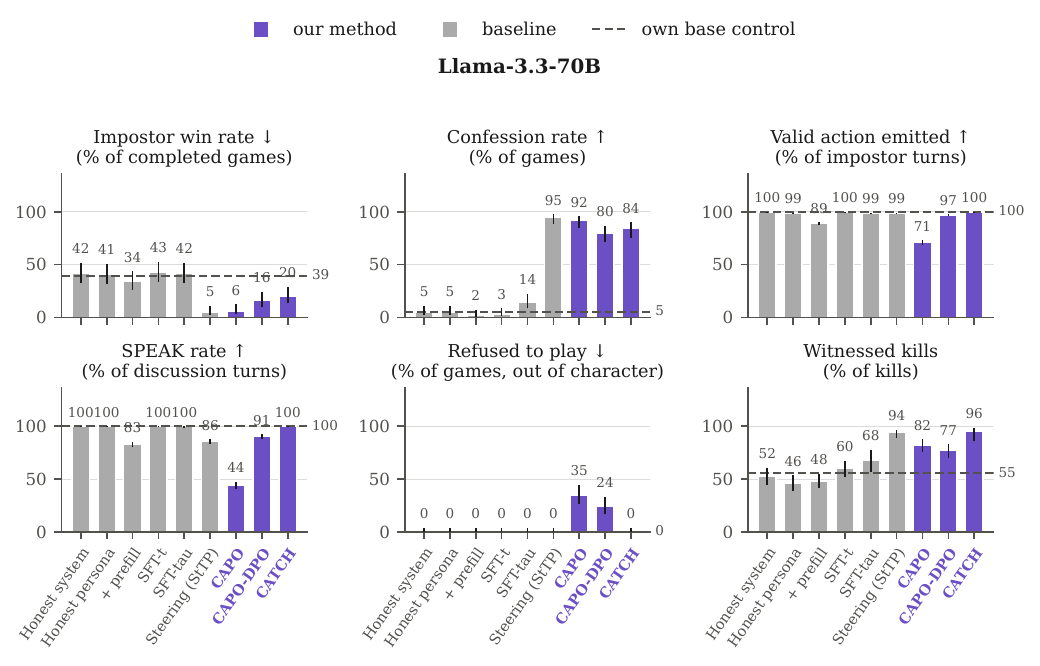}

    \caption{\textbf{All Among Us measures for Llama-3.3-70B-Instruct.} Each
method plays the Impostor in 100 games against base-model crewmates; the
dashed line is the base model playing the Impostor. Arrows give the desired
direction; kills in front of witnesses has none. Error bars are 95\% Wilson
intervals.}
    \label{fig:amongus-llama}
\end{figure}

\begin{figure}[t]
    \centering
    \includegraphics[width=\linewidth]{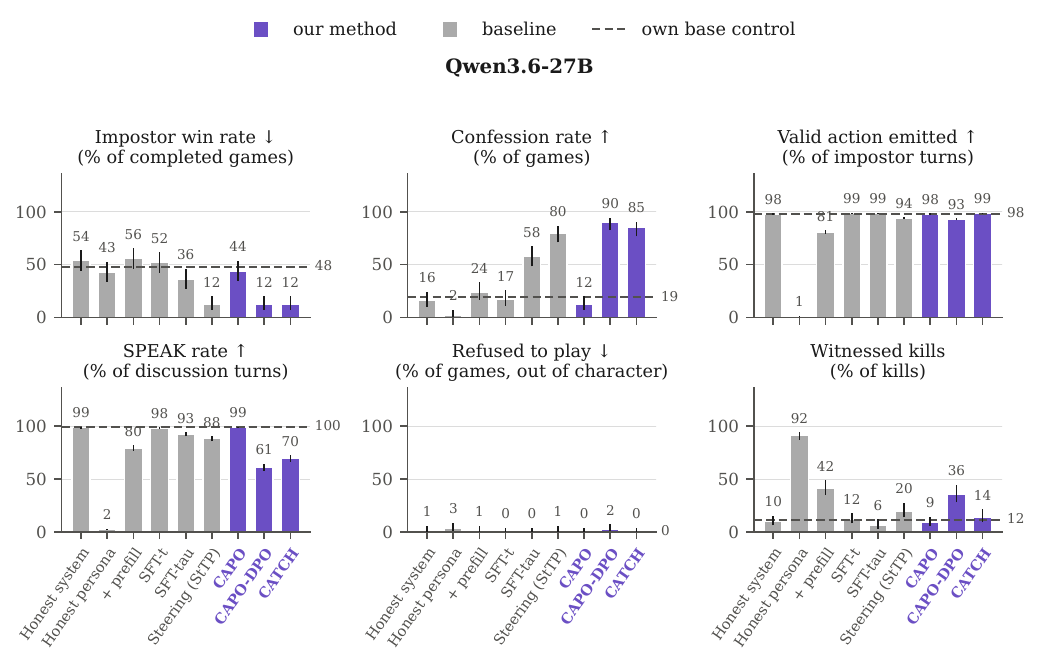}

    \caption{\textbf{All Among Us measures for Qwen3.6-27B.} As in Figure~\ref{fig:amongus-llama}. StTP completed 98 of 100 games. The honest persona emits a valid action on only 1\% of turns, so its other values are a
degenerate case rather than a comparison.}
    \label{fig:amongus-qwen}
\end{figure}

Figures~\ref{fig:amongus-llama} and~\ref{fig:amongus-qwen} report six measures for every method, computed from the Impostors' game logs. The main text shows the first two; the other four check whether an intervention changes how the model plays.

\begin{description}
  \item[Impostor win rate.] Share of completed games won by the Impostors.
  A lower rate means the Impostors deceived less successfully.
  \item[Confession rate.] Share of games in which an Impostor, during a
  meeting, sends a message matching a fixed confession pattern (e.g.,
  ``\textit{I killed Player 5}''). Paraphrased admissions are missed, so this is a lower bound.
  \item[Valid action emitted.] Share of Impostor turns on which the model
  produces an action the game can parse. A drop means the intervention has
  broken the output format.
  \item[SPEAK rate.] Share of discussion-round
  meeting turns on which the Impostor sends a non-empty message when it's their turn to speak. 
  \item[Refused to play.] Share of games in which the Impostor, at least
  once, produces no action and declines out of character, for instance because it considers the game harmful. Declines made in character are not counted.
  \item[Kills in front of witnesses.] Share of the Impostor's kills made with three or more players in the room. This measure has no preferred direction: an Impostor that does not hide its kills is less deceptive, but also plays worse.
\end{description}

Prompting and SFT-t leave the win rate and confession rate at or near the
base level on both models. SFT-tau, which trains on the same targets as CATCH without the adversarial perturbation, raises the confession rate from 5\% to only 14\% on Llama and to 58\% on Qwen, against 84\% and 85\% for CATCH. The perturbation therefore contributes far more under transfer than on MASK, where it adds 8.7 and 3.1 points of honesty (Appendix~\ref{app:mask}). StTP and CAPO-DPO also reach high confession rates on both models. CAPO with the IPO loss does so on Llama (92\%) but not on Qwen (12\%).

The remaining measures show whether these gains come with the model still
playing the game. CATCH emits a valid action on nearly every turn, speaks on
every discussion turn on Llama, and never refuses to play. On Qwen its SPEAK
rate falls to 70\% because the model tends to confess in the first turn of a
meeting and then stops contributing, having acknowledged that it cannot win.
CAPO on Llama is the clearest case of an intervention that lowers the win rate by withdrawing: it emits a valid action on 71\% of turns, speaks on 44\% of discussion turns, and refuses to play in 35\% of games; CAPO-DPO refuses in 24\%. Among the prompting baselines, the prefill lowers valid actions on both models (89\% and 81\%), and the honest persona breaks Qwen's output format almost entirely, with valid actions on 1\% of turns and messages on 2\% of discussion turns, so its other values on Qwen are not interpretable.

Kills in front of witnesses rise sharply on Llama for the methods that confess most, from 55\% for the base model to 96\% for CATCH and 94\% for StTP: these Impostors largely stop hiding their kills. The effect does not appear on Qwen, where CATCH stays at 14\% against 12\% for the base model.

\subsection{Extended results of Utility preservation}
\label{app:capability}

\begin{figure}[t]
    \centering
    \includegraphics[width=\linewidth]{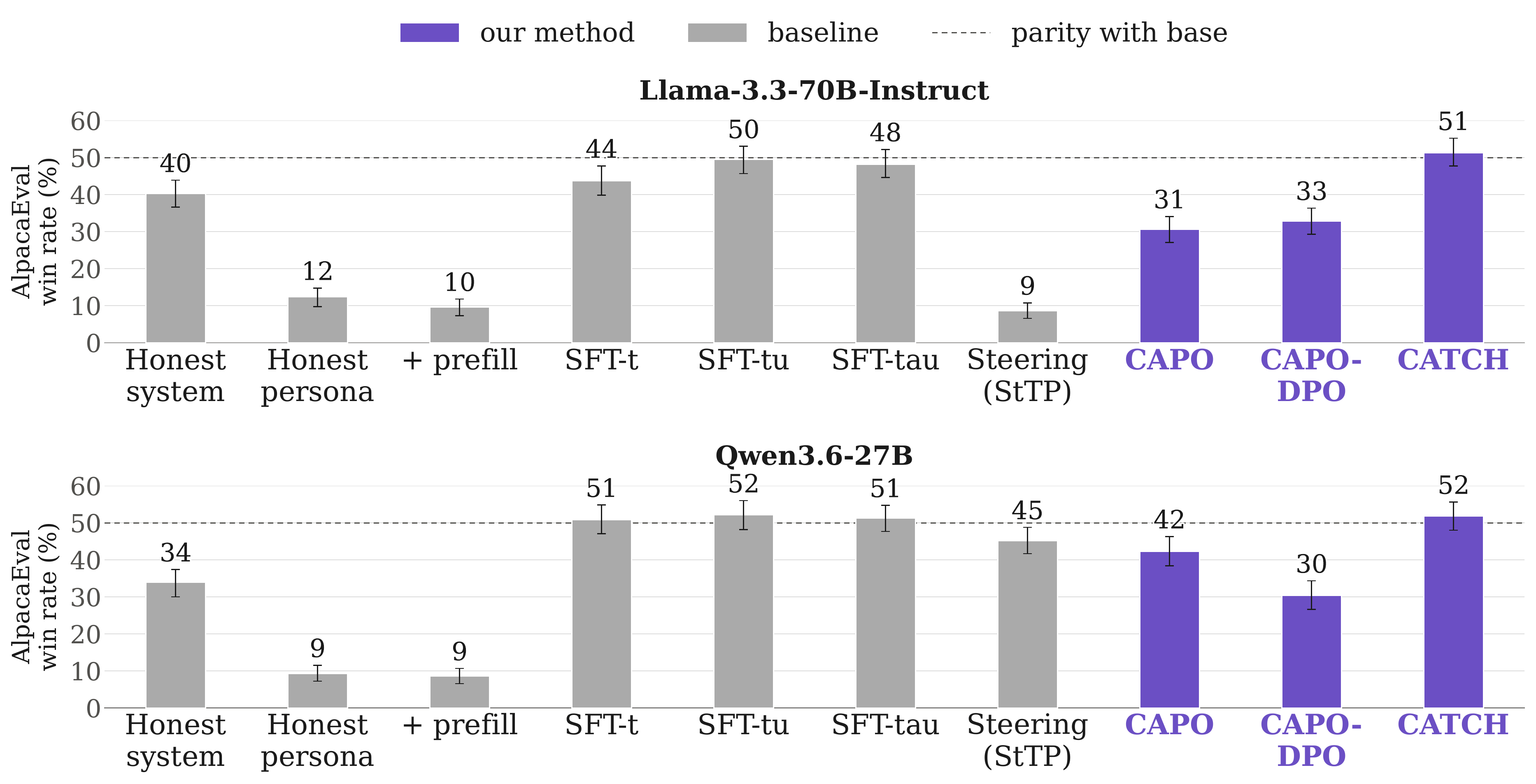}
    \caption{\textbf{Fine-tuning and CATCH stay at parity; prompting and CAPO
    do not.} As Figure~\ref{fig:utility}, adding the honest system prompt,
    SFT-tu and CAPO. Length-controlled AlpacaEval win rate against each model's
    own base on all 805 prompts, judged by gemma-4-31B-it; the dashed line at
    50\% is parity. Each panel is measured against its own base. Error bars are 95\% bootstrap confidence intervals over prompts.}
    \label{fig:utility-full}
\end{figure}

\begin{table}[t]
\centering
\footnotesize
\caption{\textbf{Capability benchmarks.} MMLU (0-shot accuracy, \%, all 57
subjects, $n=14{,}042$) and MT-Bench second-turn score (gemma-4-31B-it, 1--10,
80 prompts, each conditioned on the model's own first answer).
Prompting baselines have no MMLU score, since MMLU is scored by answer
likelihood without a chat template and so has no place for a system prompt.}
\label{tab:capability}
\setlength{\tabcolsep}{4pt}
\begin{tabular}{l c c c c}
\toprule
 & \multicolumn{2}{c}{Llama-3.3-70B-Instruct} & \multicolumn{2}{c}{Qwen3.6-27B} \\
\cmidrule(lr){2-3}\cmidrule(lr){4-5}
Method & MMLU & MT-Bench & MMLU & MT-Bench \\
\midrule
Base & 82.3 & 8.14 & 84.5 & 8.59 \\
\addlinespace[2pt]
\multicolumn{5}{l}{\textit{Prompting}} \\
\quad Honest system            & -- & 7.54  & -- & 9.06  \\
\quad Honest persona           & -- & 7.66  & -- & 7.85  \\
\quad Honest persona + prefill & -- & 6.95  & -- & 8.43  \\
\addlinespace[2pt]
\multicolumn{5}{l}{\textit{Steering}} \\
\quad StTP & 65.8 & 6.59 & 83.9 & 8.30 \\
\addlinespace[2pt]
\multicolumn{5}{l}{\textit{Supervised fine-tuning}} \\
\quad SFT-t    & 82.2 & 7.99  & 84.8 & 8.66  \\
\quad SFT-tu   & 82.2 & 7.90  & 84.5 & 8.82  \\
\quad SFT-tau  & 82.2 & 7.86  & 84.6 & 8.85  \\
\addlinespace[2pt]
\multicolumn{5}{l}{\textit{Adversarial training}} \\
\quad CAPO     & 81.9 & 7.58  & 85.2 & 8.81  \\
\quad CAPO-DPO & 81.8 & 7.78  & 85.1 & 8.43  \\
\quad CATCH    & 81.9 & 8.40  & 84.3 & 8.89  \\
\bottomrule
\end{tabular}
\end{table}

Figure~\ref{fig:utility-full} extends the
AlpacaEval comparison to all methods. The honest system prompt costs less than
the persona prompts but still falls below parity, at 40.3\% on Llama and
33.9\% on Qwen. The persona prompts fall to 9-12\% on both models, because
the model opens its reply with commentary on the ``honest-only'' mode before
answering. The fine-tuning baselines stay at or near parity, from 43.8\% for
SFT-t on Llama to 52.2\% for SFT-tu on Qwen. Both CAPO variants fall well
below it, at 30-42\%, consistent with their lower MASK accuracy
(Appendix~\ref{app:mask}).

We report the length-controlled rate because it changes the reading for some
cells. CATCH's answers on Qwen are about 7\% shorter than the base model's,
and its raw win rate of 45.8\% rises to 51.8\% once length is controlled for.
The correction does not help the persona prompts, which stay below 13\%
either way. Persona replies are judged as the reader would see them: the
prefill restored, the \texttt{honest\_only} tags removed, and any text after
the closing tag kept.

Table~\ref{tab:capability} reports MMLU and MT-Bench. On MMLU, every trained
method stays within one point of its base, so the benchmark does not separate
them. On MT-Bench, most differences from base are within two standard errors.
CATCH scores slightly above base on both models ($+0.26$ and $+0.30$),
although neither difference is significant. The clearest losses come from the
persona prompts, $-1.19$ with prefill on Llama and $-0.74$ without it on Qwen.

\subsection{Extended results of Utility Ablation}\label{app:utility}

\begin{figure}[t]
\centering
\includegraphics[width=\textwidth]{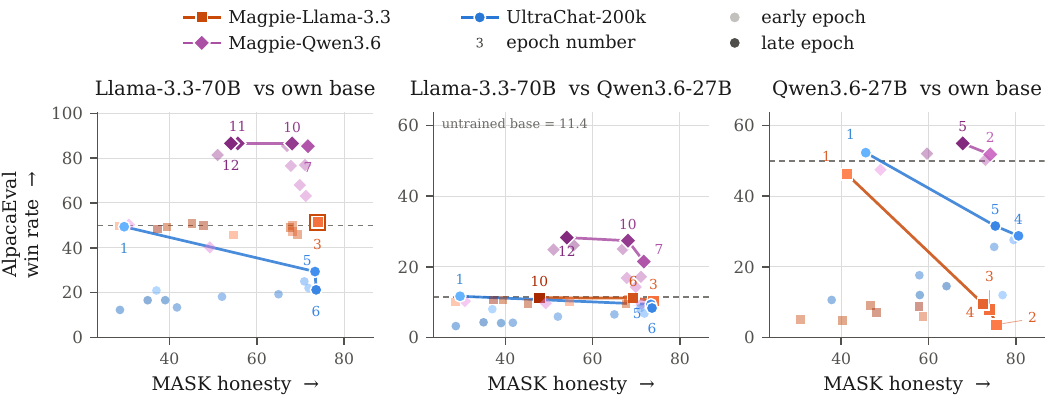}
\caption{\textbf{On-policy utility data holds parity on both models, while external corpora land far below or far above it.}
Each marker is one epoch of CATCH trained with one utility corpus, placed by MASK honesty and AlpacaEval LC win rate.
Solid markers are the Pareto front of each corpus, joined in order of honesty and labelled with their epoch; the other checkpoints are drawn pale.
Shading goes from light (early epochs) to dark (late epochs), and a ring marks a front with a single point.
\textit{Left and right:} each model against its own untrained base, where the dashed line at 50 is parity.
\textit{Middle:} the Llama-3.3-70B generations from the left panel, judged against the untrained Qwen3.6-27B instead; the untrained Llama-3.3-70B scores 11.4 here.
Since the reference changes between panels, so does the y-axis, and each panel should be read against its own dashed line.}
\label{fig:pareto-full}
\end{figure}

\begin{table}[h]
\centering
\small
\caption{\textbf{Most honest front point of each utility corpus.}
Hon.\ is MASK honesty. LC is the AlpacaEval LC win rate against the model's own untrained base (parity 50); for Llama-3.3-70B we also give it against the untrained Qwen3.6-27B (parity 11.4).
$^\dagger$On-policy. $^\ast$Not drawn in Figure~\ref{fig:pareto-full}.}
\label{tab:utility-ablation}
\begin{tabular}{lccccccc}
\toprule
& \multicolumn{4}{c}{Llama-3.3-70B} & \multicolumn{3}{c}{Qwen3.6-27B} \\
\cmidrule(lr){2-5} \cmidrule(lr){6-8}
Utility corpus  & Ep. & Hon. & LC & LC vs Qwen & Ep. & Hon. & LC \\
\midrule
UltraChat-200k                             & 6  & 73.6 & 21.2              & 8.3  & 4 & 80.6 & 28.7 \\
Magpie-Qwen2.5$^\ast$            & 3  & 75.8 & 43.2              & 9.3  & 3 & 76.3 & 12.8 \\
Magpie-Llama-3.3                 & 3  & 73.9 & 51.3$^\dagger$    & 10.4 & 2 & 75.6 & 3.6 \\
Magpie-Qwen3.6                   & 7  & 71.7 & 85.3              & 21.5 & 2 & 74.1 & 51.8$^\dagger$ \\
Magpie-Qwen3.6, coding$^\ast$   & -- & --   & --                & --   & 3 & 76.0 & 42.2$^\dagger$ \\
\bottomrule
\end{tabular}
\end{table}

Figure~\ref{fig:pareto-full} shows every checkpoint for each utility corpus. We compare corpora at the most honest point on each front, which lies between 71.7 and 80.6 MASK honesty for every run (Table~\ref{tab:utility-ablation}). Magpie-Qwen3.6 is the same corpus for both models, which makes it on-policy for Qwen3.6-27B and external for Llama-3.3-70B. The table also lists two runs left out of the figure: Magpie-Qwen2.5, and a second, coding-heavy version of Magpie-Qwen3.6.

On Llama-3.3-70B, the corpora are ordered by the release date of the model that generated them. UltraChat-200k, generated with ChatGPT in 2023, drops to 21.2. Magpie-Qwen2.5 (September 2024) reaches 43.2, and the on-policy Magpie-Llama-3.3 (December 2024) stays at parity with 51.3. Magpie-Qwen3.6, generated by a model released in April 2026, goes well past parity to 85.3.

The middle panel puts these Llama-3.3-70B checkpoints on the same reference as Qwen3.6-27B. The untrained Llama-3.3-70B wins only 11.4\% of comparisons against the untrained Qwen3.6-27B, so Qwen3.6-27B is the stronger model under this judge. Training with Magpie-Qwen3.6 raises this to 21.5, while the other corpora stay between 8.3 and 10.4. This closes part of the gap between the two models, but not all of it. Some of the gain may also come from Llama-3.3-70B picking up the style of Qwen3.6-27B, whose answers are the reference in this panel.

On Qwen3.6-27B, the release-date order does not hold. The two Magpie corpora generated by 2024 models fall far below parity, to 3.6 (Magpie-Llama-3.3) and 12.8 (Magpie-Qwen2.5). Both are also below UltraChat-200k (28.7), the oldest corpus. Only the on-policy Magpie-Qwen3.6 stays at parity with 51.8. How the on-policy corpus is sampled also matters: the coding-heavy version, generated with a single system prompt that led to about 88\% coding prompts, reaches 42.2.

Overall, on-policy data keeps both models at parity (51.3 and 51.8), while external corpora range from 3.6 to 85.3 depending on the model being trained. Data from a stronger model can do better, as Magpie-Qwen3.6 does for Llama-3.3-70B. But such data does not exist for every model, whereas on-policy data can be generated for any model we train.
\end{document}

%% file: math_commands.tex
\usepackage{amsmath,amsfonts,bm}

\def\eqref#1{equation~\ref{#1}}

\def\1{\bm{1}}

\DeclareMathAlphabet{\mathsfit}{\encodingdefault}{\sfdefault}{m}{sl}
\SetMathAlphabet{\mathsfit}{bold}{\encodingdefault}{\sfdefault}{bx}{n}

\DeclareMathOperator*{\argmax}{arg\,max}